%% file: main_final.tex
\documentclass[journal]{IEEEtran}
\usepackage{amssymb}
\usepackage{framed}
\usepackage{cite}
\usepackage{amsmath,amssymb,amsfonts}
\usepackage{graphicx}
\usepackage{textcomp}
\usepackage{xcolor}
\usepackage{cite}
\usepackage{bbold}
\usepackage{ifpdf}
\usepackage{times}
\usepackage{layout}
\usepackage{float}
\usepackage{afterpage}
\usepackage{amsmath}
\usepackage{amstext}
\usepackage{amssymb,bm}
\usepackage{latexsym}
\usepackage{color}
\usepackage{dblfloatfix}    
\usepackage{kantlipsum}    

\usepackage{graphicx}
\usepackage{amsmath}
\usepackage{amsthm}
\usepackage{graphicx}
\usepackage{pstricks}
\usepackage{subfig}
\usepackage{caption}
\usepackage{booktabs}
\usepackage{multicol}
\usepackage{lipsum}

\usepackage{enumitem}
\input{symbols.tex}
\usepackage[bookmarks,colorlinks]{hyperref}

\begin{document}
%
\title{Unfolded Algorithms for Deep Phase Retrieval}

%
%
%

\author{Naveed~Naimipour$^\star$,~\IEEEmembership{Student~Member,~IEEE,}
        Shahin~Khobahi$^\star$,~\IEEEmembership{Student~Member,~IEEE,}
        and~Mojtaba~Soltanalian,~\IEEEmembership{Senior Member,~IEEE}
\thanks{$^\star$The first two authors contributed equally to this work.}\thanks{This work was supported in part by  an Illinois Discovery Partners Institute (DPI) Seed Award. Parts of this work have been presented at the 2020 Asilomar Conference on Signals, Systems, and Computers \cite{UPR_Conf}.}
\thanks{The authors are with the Department of Electrical and Computer Engineering, University of Illinois at Chicago, Chicago, IL 60607, USA. Corresponding Author: Naveed Naimipour (e-mail: nnaimi2@uic.edu).}}


%



\maketitle

\begin{abstract}
Exploring the idea of phase retrieval has been intriguing researchers for decades, due to its appearance in a wide range of applications. The task of a phase retrieval algorithm is typically to recover a signal from linear phase-less measurements. In this paper, we approach the problem by proposing a hybrid model-based data-driven deep architecture, referred to as \emph{U}nfolded \emph{P}hase \emph{R}etrieval (\emph{UPR}), that exhibits significant potential in improving the performance of state-of-the-art data-driven and model-based phase retrieval algorithms. The proposed method benefits from versatility and interpretability of well-established model-based algorithms, while simultaneously benefiting from the expressive power of deep neural networks. In particular, our proposed model-based deep architecture is applied to the conventional phase retrieval problem (via the incremental reshaped Wirtinger flow algorithm) and the sparse phase retrieval problem (via the sparse truncated amplitude flow algorithm), showing immense promise in both cases. Furthermore, we consider a joint design of the sensing matrix and the signal processing algorithm and utilize the deep unfolding technique in the process. Our numerical results illustrate the effectiveness of such hybrid model-based and data-driven frameworks and showcase the untapped potential of data-aided methodologies to enhance the existing phase retrieval algorithms.

\end{abstract}


%
\IEEEpeerreviewmaketitle

\section{Introduction}
The task of phase retrieval is concerned with recovering a complex or real-valued signal of interest, $\bx\in\mathds{R}^{n}/\mathds{C}^{n}$, from $m$ linear phase-less measurements of the form
\begin{eqnarray}
\label{eq:acq}
y_i = |\langle\ba_i, \bx\rangle|,\,\, \mathrm{for}\,\,i  \in \{ 1, 2, \cdots, m\},
\end{eqnarray}
where the set of sensing vectors $\{\ba_i \in \mathds{R}^{n}/\mathds{C}^{n}\}_{i=1}^{m}$, are assumed to be known \emph{a priori}. The journey of solving the decades old phase retrieval problem has led to numerous algorithms and methodologies. This is no surprise given the many applications of phase retrieval, including those in areas such as crystallography, optics, and imaging \cite{millane,kim}. These extensive practical applications have made their way into deep space as well, where phase retrieval plays an important role in signal processing for space telescopes such as NASA's Hubble Space Telescope \cite{Fienup93, Krist95, Sarnik}. With the growing number of applications in various fields, the developed methodologies continue to increase in number and efficiency. 
Note that a large number of methods in the literature have their roots in the seminal works by Gerchberg, Saxton, and Fienup \cite{GS1,GS2,fienup1,fienup2,fienup3}. Their extensive body of work was integral to the implementation of phase retrieval algorithms in NASA's Hubble Space Telescope \cite{Fienup93,Krist95}, and is still widely referenced in modern phase retrieval research. However, the Gerchberg-Saxton algorithm's shortcomings in terms of finding the optimal solution in an efficient manner has resulted in numerous new directions.  While some convex formulations have been proposed in the literature \cite{candes2013phaselift, candes2014solving,jaganathan2013sparse}, most of the existing phase retrieval algorithms view the problem through a non-convex lens. \nocite{liang2017phase,qiu2016prime}
As a case in point, methodologies such as Wirtinger flow (WF), truncated Wirtinger flow (TWF),  reshaped Wirtinger flow (RWF/IRWF), and incremental truncated Wirtinger flow (ITWF) have all shown promise in addressing the problem in an efficient and accurate manner \cite{candes,chen,TWF,RWF}.   WF was the first approach to display the significance of spectral initialization and updating. Since then, other approaches such as RWF/IRWF improved the performance further by implementing their own spectral initialization and updating method \cite{candes,RWF}. 

In many practical applications, the signal of interest is naturally sparse or can be made sparse by design  \cite{sparseseen, SPARTA}. This has resulted in approaches that reduce the number of required  measurements  via block-sparse phase retrieval solvers, such as the TWF \cite{TWF}. In addition, more robust approaches such as sparse truncated amplitude flow (SPARTA) have been formulated and shown to further improve performance \cite{SPARTA}. SPARTA's two-stage approach of simple power iterations for initializations, and truncated gradient calculations via thresholds, makes it a particularly interesting. Despite the enormous progress made, there remains a number of obstacles including the reduced applicability of algorithms developed based on the Gaussian noise assumption for other kinds of disturbances \cite{AltGD}. Particularly, developing methods that can deal with outliers is especially important. 
In this context, algorithms such as AltIRLS and AltGD \cite{AltGD} can handle the existence of impulsive noise using an $\ell_p$-fitting based estimator. Such algorithms are of particular interest due to their ability to outperform versatile algorithms such as TWF. Moreover, most established phase retrieval algorithms struggle with internal parameter optimization, such as determining the optimal step size for signal recovery while avoiding to fall into local minima when sample sizes are small. This is still the case even with the improved algorithms such as RWF \cite{RWF}. In fact, computational inefficiency is still a major obstacle in applying such phase retrieval algorithms in large-scale or real-time scenarios.


Along with the convex and non-convex divide, the phase retrieval algorithms can also be categorized based on whether they are model-based or data-driven: model-based methodologies, like those discussed above, seek to design algorithms  through a preliminary modeling of the problem and incorporating the reasoning that emerge from the model, whereas data-driven methodologies rely primarily on data to solidify the workings of algorithms. 
Data driven approaches can help with some of the shortcomings mentioned earlier 
by making use of  the expressive power of deep neural networks and 
training them in a manner that the resulting network acts as an estimator of the true signal given the measurements vector $\by= |\bA^H\bx|$. Within the realm of phase retrieval, deep learning has been only used to design neural nets for  algorithms such as hybrid-input-output (HIO) and Fienup's method \cite{IntroDNN,Isil}. Such works are limited due to their inability to deal with multiple types of models, as well as the more recent phase retrieval algorithms that are more complex in nature. Additional body work has been done using convolutional neural nets, such as prDeep \cite{metzler}, leading to a separate class of architectures not versatile enough to improve the existing algorithms. In fact, some of Fienup's recent work for space telescopes take advantage of smart initializations obtained from convolutional neural nets for phase retrieval \cite{FienupML}. The said limitations directly relate to  the two major pitfalls of data-driven approaches, i.e., (i) their need to a relatively large amount of data for training purposes and (ii) their inherent lack of interpretability, even after training. Still, data-driven approaches, such as deep learning techniques, have become immensely useful in recent years for handling complex signal processing problems. As a result, it is not difficult to see why a hybrid model-based and data-driven model has the potential to improve the data acquisition model even further by effectively dealing with each approach's weaknesses. For complex systems, such a hybrid model is promising due to its ability to integrate parameterized and non-parameterized mathematical models. In addition, the employment of particular activation functions for model-based deep architectures allows them to be differentiated from conventional deep learning methodologies. Specifically, the activation functions can be designed to mimic non-linearities already present in traditional optimization algorithms.

The \emph{deep unfolding} technique is an effective amalgam of model-based and data-driven approaches. LeCun et al. first introduced this notion in \cite{lecun}; which was extended thoroughly by Hershey et al. in \cite{hershey2014deep}. Deep unfolding facilitates the design of model-aware deep architectures based on well-established iterative signal processing techniques. Deep unfolded networks have repeatedly displayed great promise in the field of signal processing and can make use of the immense amounts of data, along with the domain knowledge gleaned from the underlying problem at hand \cite{farsad2020data,bertocchi2019deep,khobahi2019deep,shlezinger2020deepsic,khobahi2019model,balatsoukas2019deep,shlezinger2019viterbinet,agarwal2020deep}\nocite{khobahi2020deep}. Furthermore, they have the ability to utilize the adaptability and reliability of model-based methods, while also taking advantage of the expressive power of deep neural networks. As a result, they are an ideal candidate for problems such as phase retrieval, particularly in non-convex settings where researchers  struggle with bounding the complexity of signal processing algorithms while keeping them effective. This problem particularly arises when applying iterative optimization techniques. In this context, first-order methods are widely used as iterative optimization techniques with low \emph{per-iteration} complexity. 
The drawback, on the other hand, stems from their tendency to suffer from a slow speed of convergence, since they usually require many iterations to converge. It should also be noted that predicting the number of iterations required for convergence is generally a difficult task. As a result, they may not be recommended for low-latency and reliable signal processing in real-time applications. It is therefore logical to consider fixing the total number of iterations of such algorithms (say $L$ iterations), while at the same time, seeking to optimize the (parameters of the) iterations in order to achieve the best improvement in the underlying objective function at hand. Accordingly, our goal in this paper is to improve the existing first-order methods by \emph{meta optimizing} the SPARTA and IRWF iterations when the total computational budget is fixed. This is done by formulating the meta-optimization problem in a deep learning setting, and interpreting the resulting unrolled iterations as a neural network with $L$ layers where each layer is designed to imitate one iteration of the original optimization method. Eventually, such a deep neural network can be trained using a small data-set and the resulting network can be used as an enhanced first-order method for solving the underlying problem at hand. 



A particular concern with regards to  
phase retrieval is the usual  
assumption that the sensing matrix is known \emph{a priori}. In the existing literature, designing the phase retrieval algorithms has received the primary attention. Designing the measurement matrix, on the other hand, is considerably unique. Random matrices have been shown to be effective in related areas such as compressive sensing, but typically need to satisfy certain criteria to guarantee recovery. A particularly well-researched criterion that can guarantee said recovery is the Restricted Isometry Property (RIP) \cite{candesRANDOM1}, which is known to be a sufficient condition for noise-robust sparse signal recovery \cite{candesRANDOM2}. Alternatively, deterministic design of the sensing matrix presents a set of advantages that can counteract the pitfalls that stem from the random selection of matrices. Although the RIP is a well-known criterion for random matrices, testing the matrices for RIP compliance may not be efficient from a computational viewpoint. Furthermore, such random matrices themselves tend to become large and unmanageable when the signal size grows large. Such issues with random matrices naturally translate to the design of measurement matrices in phase retrieval as well \cite{candes,randCAND, randCAND2}, but the corresponding solutions have not been looked into thoroughly. Deterministic designs allow for judicious matrix constructions that can mitigate the efficiency and storage issues that are associated with random matrix designs. In addition, they can account for the underlying distribution of the signal and system parameters, which is the main shortcoming of a typical random matrix approach. As such, an architecture that could design the sensing vectors/matrices in a deterministic setup would be extremely valuable in practical scenarios \cite{liGE}. Whether it is astronomy, X-ray imaging, or any other application, designing a sensing matrix for each signal is a difficult task \cite{xrayCCD, Fienupastro}. 

\nocite{adaptiveSparse, adaptive2}

We note that there already have been a number of research works on designing the sensing matrix, including designs based on learning methodologies \cite{designNorm, adaptiveSparse, adaptive2, designCNN, designCNN2, designMMV, designSDA}. The interest in  learning techniques such as convolutional neural nets and stacked denoised autoencoders \cite{designCNN, designCNN2, designSDA} has increased greatly. This has led to works such as \cite{8055634}, where the measurement matrix is designed based on mutual information, but the ideas are not  exploited for recovery. A deep architecture that could handle the design of the sensing matrix along with the recovery task would be able improve the accuracy and efficiency of phase retrieval in a way unmatched by previous methods. As result, we undertake a deterministic task-specific and data-specific design of the sensing matrix that can be cascaded to the recovery algorithm, resulting in an immense improvement in the recovery performance.

\vspace{3pt}

\textbf{Contributions Overview:} In this paper, we propose model-aware deep architectures, referred to as Unfolded Phase Retrieval (UPR)---as a new approach to the problem of phase retrieval. In particular, we focus on two scenarios: the conventional phase retrieval problem (with IRWF) and the sparse phase retrieval problem (with SPARTA), which have both shown great promise. Furthermore, we propose a general framework for designing task-specific sensing matrices for improving the performance of the underlying recovery algorithm and to outperform numerous state-of-the-art algorithms as a supplementary feature. Deep unfolding has the unique distinction of falling into the category of a hybrid model-based and data-driven methodology. More explicitly, we consider a joint design of the sensing matrix and the signal processing algorithm and utilize the deep unfolding technique in the process. As a result, our hybrid methodology can adopt the advantages of both methodologies as well. Specifically, we have the capability to exploit the data for better accuracy and performance, the resulting interpretability leads to trusted outcomes, we require less data due to having less parameters to train, and obtain an enhanced convergence rate. UPR allows for a unification of the optimization process for system parameters in an end-to-end manner. 
Interestingly, UPR acts as a unified framework for both optimal sensing and recovery. As mentioned before, such a framework can be cascaded to the recovery algorithm which allows for noticeable recovery accuracy improvement. Additionally, UPR has a remarkable compatibility with gradient based algorithms, which is displayed both in our description of the methodology and in our implementation with SPARTA and IRWF. Finally, we compare our results with the said baseline algorithms and different variants of the RWF algorithm that have already shown good performance in the context of phase retrieval. 
We show that UPR significantly outperforms the baseline algorithms from which it has emerged.


\vspace{3pt}

\textbf{Organization of the Paper:} 
Section II is devoted to a technical overview of the system model and a brief  overview of deep unfolding in the context of phase retrieval. Section III  introduces the UPR framework, its baseline algorithms and its initialization method, as well as our  methodology for the design of the sensing matrix. We examine the performance of UPR in Section IV. Finally, Section V concludes the paper.  

\vspace{3pt}

\textbf{Notations:} Throughout the paper, we use boldface lowercase letters to denote vectors, and boldface capital letters to denote matrices. Calligraphic letters are reserved for sets.  The superscripts $(\cdot)^*$, $(\cdot)^T$, and $(\cdot)^H$ represent the conjugate, the transpose, and the Hermitian operators, respectively. $\| \cdot \|_2$ represents the Euclidean norm of the vector argument. The operator symbol $ \odot $ represents the Hadamard product of matrices.  

\section{System Model and Problem Formulation}
\label{sec:model}

In this section, we present a mathematical formulation of the phase retrieval problem and provide a brief overview of the existing classical model-based phase retrieval algorithms which will lay the ground for the proposed hybrid model-based and data-driven methodology.

As discussed earlier, a phase retrieval \emph{system} can be mathematically formalized by considering the following encoding module (i.e., the data-acquisition system):
\begin{align}\label{eq:sysmodel}
    \text{Encoding Module:}\quad \by =  \left|\bA\bx \right|,
\end{align}
where $\bx\in\mathds{C}^n$ denotes the underlying signal of interest, $\bA = [\ba_1, \ba_2, \cdots, \ba_m]\in\mathds{C}^{n\times m}$ denotes the sensing matrix, and $\by\in\mathds{R}^m$ represents the captured phase-less measurements of the signal $\bx$.

Traditionally, in a phase retrieval problem we seek to retrieve the measurement phase by recovering the signal of interest $\bx$ from the embedded phase-less measurements $\by$ given the knowledge of the sensing matrix $\bA$. Specifically, the phase retrieval problem under the noise-free assumption can be formally stated as:
\begin{align}\label{eq:pr}
    \mathcal{P}_0:\quad&\mathrm{find}\quad \bx\quad \mathrm{s.t.} \quad \by = \left|\bA\bx\right|,
\end{align}
where the constraint in $\mathcal{P}_0$ represents the feasible set of the problem corresponding to the data consistency principle. Clearly, the above program is non-convex due to its non-convex constraint. Moreover, observe that if $\bx^{\dagger}$ is a feasible point of the problem $\mathcal{P}_0$, then $e^{-j\phi}\bx^{\dagger}$ is also a solution to the problem for an arbitrary phase constant $\phi$. Hence, it is only possible to recover the underlying signal of interest up to a phase shift factor in a phase retrieval problem. With that in mind, let $\bx^{\dagger}$ be any solution to the phase retrieval problem defined in \eqref{eq:pr} and let $\bx^{\star}$ represent the true  signal of interest. A meaningful quantifying metric of closeness of the recovered signal $\bx^{\star}$ to the true signal can then be defined as follows:
\begin{align}\label{eq:metric}
    \mathrm{D}(\bx^{\star}, \bx^{\dagger}) = \underset{\phi\in[0,2\pi)}{\mathrm{min}}\|\bx^{\star} - e^{-j\phi}\bx^{\dagger}\|_2^2,
\end{align}
which was first proposed in \cite{candes} and can be interpreted as measuring the Eucledian distance of two complex vectors up to a global phase difference. 

In this work, we consider the phase retrieval problem from the perspectives of \emph{system design} and \emph{reconstructive algorithm development}. 
A system design approach for a phase retrieval problem is mainly concerned with finding the best set of parameters of the encoder module that facilitate the recovery of the underlying signal of interest from the phase-less measurements captured through the encoder module (i.e., the data acquisition system). In particular, a system design perspective for a phase retrieval data acquisition system can be defined as designing the sensing matrix in a deterministic fashion according to a performance criterion. As an example, \cite{8055634} has considered the design of the sensing matrix in a deterministic manner such that it maximizes the mutual information between the signal of interest and the acquired phase-less measurements. However, the development of effective signal reconstruction techniques that can harness this maximal mutual information obtained by a judicious design of the sensing matrix is still an open problem. Hence, a more natural approach to this problem is to consider a joint design of the sensing matrix and the recovery algorithm.

We seek to jointly design the sensing matrix (encoder module) and the reconstruction algorithm (decoder module), in contrast to the existing methodologies that either consider the development of the reconstruction algorithm or the design of the sensing matrix. The proposed methodology can be viewed as a unification of both approaches. Specifically, we propose a hybrid model-based and data-driven methodology to this problem and utilize the notions of auto-encoders and the deep unfolding technique as the main tools to develop our proposed framework. By implementing such a methodology, while the sensing matrix will be designed to maximize the signal reconstruction accuracy at the decoder module, the decoder module itself will be trained to perform the recovery by accounting for the learned sensing matrix---leading to the promised joint design. 


In order to make the joint design possible, we seek to identify sensing matrices that make the recovery of the underlying signal of interest feasible and to identify structures that are amenable to a specific class of reconstruction algorithms and vice versa. Let $\mathcal{D}_{\Phi}: \mathds{R}^n \mapsto \mathds{C}^m$ denote the class of decoder functions parameterized on a set of parameters $\Phi$. Further assume that for a given measurement matrix $\bA$ and the corresponding measurements vector $\by$, $\hat{\bx}$ represents an estimation of the signal of interest provided by a certain characterization of the decoder function, viz.
\begin{align}\label{eq:estimation}
    \hat{\bx} = \mathcal{D}_{\phi}\left(\by; \bA\right), 
\end{align}
where $\phi\in\Phi$ denotes a characterization of the decoder module from the original class.
Recall that for a fixed sensing matrix $\bA$, the dynamics of the data-acquisition model of a phase retrieval system is given by $\by = |\bA\bx|$. Accordingly, for a fixed characterization of the decoder function $\mathcal{D}_{\phi\in\Phi}$, the design of the sensing matrix $\bA \in \mathds{C}^{m\times n}$  can be considered according to the following optimization problem:
\begin{align} \label{eq:A}
    \underset{\bA\in\mathds{C}^{m\times n}}{\mathrm{min}} \;\;\mathrm{E}_{\bx\sim P(\bx)}\left\{\mathrm{D}\left(\bx, \mathcal{D}_{\phi}(|\bA\bx|; \bA)\right)\right\},
\end{align}
where $P(\bx)$ represents the distribution of the underlying signal of interest. For simplicity, we drop the notation $\bx\sim P(\bx)$, and unless mentioned otherwise, all expectations are taken with respect to the input distribution in the sequel. The aforementioned approach for the design of the sensing matrix considers the signal reconstruction accuracy as a design criterion with respect to a particular realization of the decoder functions. Thus, the obtained matrix $\bA$ can be viewed as a task-specific encoding matrix which not only considers the underlying distribution of the signal, but also the considered class of decoder functions resulting in a superior performance. Once a solution to \eqref{eq:A} is obtained, the sensing matrix can be utilized for data-acquisition purposes while the fixed decoder module $\mathcal{D}_{\phi}$ can be used to carry out the reconstruction of the signal of interest. However, starting our design from \eqref{eq:A} presents us with an inherent performance bottleneck. 
To see why, let $\mathcal{D}_{\phi_1\in\Phi}$ and $\mathcal{D}_{\phi_2\in\Phi}$ represent two realizations of the decoder module from the same class, 
and let $\bA$ denote a fixed sensing matrix that will be the solution to \eqref{eq:A} for $\phi=\phi_2$. Then, for a signal of interest $\bx\sim P(\bx)$ and the corresponding phase-less measurements $\by$, it might be the case that 
\begin{align}
    \mathds{E}\left\{\mathrm{D}(\bx, \mathcal{D}_{\phi_1}(\by; \bA))\right\} \leq \mathds{E}\left\{\mathrm{D}(\bx, \mathcal{D}_{\phi_2}(\by;\bA))\right\}.
\end{align}
In other words, the realization of the decoder module on the set of parameters $\phi_1\in\Phi$, results in a more accurate reconstruction algorithm as compared to the alternative realization. As a result, it is better to consider the design of the sensing matrix with respect to $\mathcal{D}_{\phi_1}$ in such a case. Nonetheless, finding an optimal $\phi^{\star} \in \Phi$ such that 
\[
\mathds{E}\left\{\mathrm{D}(\bx, \mathcal{D}_{\phi^{\star}}(\by; \bA))\right\} \leq \mathds{E}\left\{\mathrm{D}(\bx, \mathcal{D}_{\phi}(\by;\bA))\right\}, \quad \forall \phi\in\Phi,
\]
is a particularly challenging task due to the non-convex nature of the problem. If such a characterization $\phi^{\star}$ is known, the design of the sensing matrix matrix can easily be carried out. The above observation is further evidence for the paramount importance of developing a unified framework that allows for a joint optimization of the reconstruction algorithm (i.e. finding an optimal or sub-optimal $\phi^{\star}$) and the sensing matrix. 
In the following, we consider the joint design of the sensing matrix and the signal processing algorithm and propose a novel framework based on the deep unfolding technique that allows for an optimization over the set of both design parameters $\{\Phi, \bA\}$. The resulting framework can be viewed as a hybrid data-driven model-based deep neural network that achieves a far superior performance when compared to the existing model-based and data-driven methodologies. Furthermore, the obtained deep architecture is interpretable due to the incorporation of the domain knowledge and assumes far less parameters when compared to existing black-box neural networks. Due to the fewer training parameters, the proposed architecture requires far fewer training samples for optimization of the network when compared to the existing data-driven methodologies. We will provide a high-level description of the proposed methodology and proceed deeper into the details of each module in the next section.

We conclude this section by mathematically formalizing the problem at hand. We cast the problem of jointly designing the sensing matrix and the reconstruction algorithm as the following optimization problem:
\begin{align}\label{eq:paper}
    \underset{\{\bA\in\mathds{C}^{m\times n}, \phi\in\bPhi\}}{\mathrm{min}}\;\;\mathds{E}\{\mathrm{D}\left(\bx, \mathcal{D}_{\phi}\left(|\bA\bx|; \bA\right)\right)\}.
\end{align}
Our preliminary step is to mathematically formulate a proper parameterized class of the decoder and encoder modules that facilitate the incorporation of domain knowledge. 
To this end, we consider the realization of the decoder module by first formulating the problem of phase retrieval as an optimization problem, and then, we resort to first-order optimization techniques that lay the groundwork for obtaining a rich class of parameterized decoder functions. Once such a class is formalized, we make the connection between deep neural networks and first-order optimization techniques and propose two novel model-based deep architectures that encode the domain knowledge in their respective architecture and task-specific computations. Next, we show how to tackle the above program using the prominent deep learning techniques by taking advantage of the available data. 

\section{UPR: The Proposed Framework}
In this section, we present the proposed hybrid data-driven and model-based \textbf{\emph{U}}nfolded \textbf{\emph{P}}hase \textbf{\emph{R}}etrieval (UPR) framework. 
We begin our presentation by mathematically defining the encoder and decoder modules which take the form of a model-based deep architecture, highly tailored to the problem of phase retrieval. Then, we interpret the overall dynamics of the proposed methodology as a single auto-encoder module whose training procedure takes the form of \eqref{eq:paper}, which can then be tackled using the well-established deep learning techniques such as back-propagation. Once the training of the auto-encoder module is complete, the optimized sensing matrix can be extracted from the encoder module for data-acquisition purposes, while the optimized decoder module can be used as an enhanced signal reconstruction algorithm to carry out the signal estimation task from the phase-less measurements obtained through the optimized sensing matrix.
\subsection{Architecture of the Encoder Module}
We take this opportunity to formally define the class of encoder functions for the phase retrieval problem. 
Recall that the governing dynamics of the data-acquisition system in a phase retrieval problem is given by
\begin{align}\label{eq:dac}
    \by = |\bA\bx|,
\end{align}
where the measurement vector $\by$ and the sensing matrix $\bA$ are used as the input to a signal reconstruction algorithm to obtain $\hat{\bx}$. Now, we use the above dynamics as a blue-print to design our encoder module parametrized on the sensing matrix. To this end, we define the hypothesis class $\mathcal{H}_{e}$ of the possible encoder functions (module) as follows:
\begin{align}
    \mathcal{H}_{e} = \{f_{\bA}: \bx\mapsto |\bA\bx|\,:  \bA \in \mathds{C}^{m\times n}\},
\end{align}
where w $f_{\bA}(\bx) \triangleq |\bA\bx|$. The above hypothesis class encapsulates the possible encoder modules whose computation dynamics mimics the behaviour of the data-acquisition model in a phase retrieval model. Alternatively, the above hypothesis class can be interpreted as a \emph{one-layer neural network} with $n$ input neurons and $m$ output neurons, where the activation function is given by $|\cdot|$ with the trainable weight matrix $\bA$. Specifically, the input to such a neural network is $\bx\in\mathds{C}^{n}$, while the output is  $|\bA\bx|\in\mathds{R}^m$. This paves the way for viewing the design of the sensing matrix from a statistical learning theory perspective. For instance, the problem of designing the sensing matrix for a given realization of a decoder function $\mathcal{D}_{\phi}$ with respect to the hypothesis class $\mathcal{H}_e$ can be re-expressed as the following \emph{learning problem}:
\begin{align}\label{eq:encoderlearning}
    \underset{\mathcal{E}_{\bA}\in\mathcal{H}_e}{\mathrm{min}}\;\;\mathds{E}\{\mathrm{D}\left(\bx, \mathcal{D}_{\phi}\circ\mathcal{E}_{\bA}(\bx)\right)\}.
\end{align}
The above learning problem seeks to find an encoder function $\mathcal{E}_{\bA}$ such that the resulting functions maximizes the reconstruction accuracy according to the decoder function $\mathcal{D}_{\phi}$. Equivalently,  \eqref{eq:encoderlearning} can be expressed as learning the weights of a one-layer neural network $\mathcal{E}_{\bA}(\bx) = |\bA\bx|$, where for any choice of $\bA$ we have  $\mathcal{E}_{\bA}\in\mathcal{H}_e$. Then, the problem in \eqref{eq:encoderlearning} can be re-expressed as a deep learning problem by considering the training of $\mathcal{E}_{\bA}$ according to a loss function that is the same as objective function in \eqref{eq:encoderlearning}, i.e.,
\begin{align}
    \underset{\bA\in\mathds{C}^{m\times n}}{\mathrm{min}}\;\; \mathds{E}\left\{\mathrm{D}(\bx, \mathcal{D}_{\phi}\circ\mathcal{E}_{\bA}(\bx))\right\}.
\end{align}
Hereafter, we seek to formulate the joint design of the sensing matrix and the reconstruction algorithm (i.e., see \eqref{eq:paper}) from a deep learning and statistical learning point-of-view, and eventually, formulate \eqref{eq:paper} as a purely deep learning problem and tackle it by utilizing the well-established deep learning techniques. The following subsection, in particular, is devoted to mathematically defining two parameterized classes of decoder functions; one specialized for sparse phase retrieval, and the other, specialized for the conventional class of phase retrieval problems. Moreover, we will formally introduce the UPR framework and its training procedure.

\subsection{Architecture of the Decoding Module}
In this subsection, 
we propose two model-based deep architectures that are not only highly-tailored to the problem of phase retrieval but also allow for learning task-specific superior reconstruction algorithms. 
Next, we cascade the decoder and the encoder module to obtain an auto-encoder deep architecture that will facilitate the joint training of the sensing matrix and the reconstruction algorithm.

\vspace{0.15cm}
\noindent
\textbf{\emph{Deep Unfolded Networks for Phase Retrieval.}}
We begin with 
a brief introduction to deep unfolded networks and their relation with mathematical optimization. 
Note that the goal of a decoder module is to obtain an estimate of the underlying signal of interest from the phase-less measurements of the form $\mathcal{E}_{\bA}(\bx) = |\bA\bx|$, for a given sensing matrix $\bA$. Equivalently, a decoder function seeks to tackle the following problem:
\begin{align}
\label{eq:pr1}
    \mathrm{find}\quad \bz\quad \mathrm{s.t.} \quad \mathcal{E}_{\bA}(\bz) = \mathcal{E}_{\bA}(\bx).
\end{align}
In lieu of directly tackling \eqref{eq:pr1}, a more practical approach is a reformulation of \eqref{eq:pr1} as an optimization problem to approximate of the true signal of interest. Accordingly, \eqref{eq:pr1} can be reformulated as:
\begin{align}\label{eq:pr2}
    \underset{\bz\in\mathds{C}^{n}}{\mathrm{min}}\quad \ell(\bz)\quad\mathrm{s.t.}\quad \bz\in\Omega,
\end{align}
where $\Omega$ represents the search space of the underlying signal of interest. For instance, if it is known \emph{a priori} that the signal of interest is $s$-sparse, this information can be encoded in $\Omega$, i.e., $\Omega = \{\bx\in\mathds{C}^n : \, \|\bx\|_0 = s\}$. On the other hand, for a conventional phase retrieval problem where no prior structural information is assumed on $\bx$, we set $\Omega=\mathds{C}^{m}$.

The ultimate formalization of a phase retrieval problem, as we move from \eqref{eq:pr1} to \eqref{eq:pr2}, boils down to a proper realization of the loss function $\ell(\bx)$ and the feasible set $\Omega$. In particular, for a given signal of interest $\bx$ and sensing matrix $\bA$,  if $\bz^{\star}\in\mathrm{argmin}_{\bz\in\Omega}\;\ell(\bz)$ then we must have  $\mathcal{E}_{\bA}(\bz^{\star}) = \mathcal{E}_{\bA}(\bx)$. 
Note that different choices of the loss function $\ell(\bx)$ and the feasible set $\Omega$ give rise to various flavors of recovery algorithms for a phase retrieval, as has been previously discussed. In this work, we consider the development of the class of decoder functions $\mathcal{H}_d$ such that for any realization of the decoder function $D_{\phi}\in\mathcal{H}_d$, the function $\mathcal{D}_{\phi}$ becomes a good approximator of the solution to \eqref{eq:pr2}. Note that there exist alternative approaches to tackling the phase retrieval problem. For example, the direct development of a class of decoder functions can be considered which are model-agnostic. One particular class of such model-agnostic methodologies can be derived by utilizing black-box deep neural networks. Assume that the set $\{\bx_{i}, \mathcal{E}_{\bA}(\bx_i)\}_{i=0}^{B-1}$ represents our available data from the phase retrieval system, i.e., we have $B$ realizations of the input-output relationship of our phase retrieval system for a given $\bA$. Furthermore, let 
\begin{align}\label{eq:DNN}
\mathcal{F}_{\sigma, L}=\{f_{\Gamma}:\mathcal{E}_{\bA}(\bx) \mapsto \bx\,| f_{\Gamma}(\bx) = \sigma^L_{\gamma_L}\circ \cdots \circ \sigma^1_{\gamma_1}(\mathcal{E}_{\bA}(\bx))\}
\end{align}
represent the class of all $L$-layer deep neural networks with activation function $\sigma(\cdot)$, where $\gamma_i$ denotes the weights of the $i$-th layer, and $\Gamma = \{\gamma_i\}_{i=1}^{L}$. Then, a decoder function $\hat{f}_{\Gamma^{\star}}\in\mathcal{F}_{\sigma, L}$ can be found by training the neural network; 
e.g., the realization of a decoder function with respect to some fixed sensing matrix $\bA$ is given by:
\begin{align}
    \hat{f}_{\Gamma^{\star}} &= \underset{f_{\Gamma}\in\mathcal{F}_{\sigma,L}}{\mathrm{argmin}}\;\frac{1}{B}\sum_{i=0}^{B-1}\|\bx_i - f_{\Gamma}(\mathcal{E}_{\bA}(\bx_i))\|_2^2.
\end{align}
The above model-agnostic decoder function is only justifiable in scenarios where the input-output relationship of the system is unknown (i.e. in applications such as computer vision and natural language processing problems). Furthermore, due to the model-agnostic nature of the above methodology, the obtained decoder function is not interpretable and acts as a black-box decoder in which the decision making process of the decoder function cannot be examined by the user. In addition, the above black-box methodology for solving the inverse problem results in a very large number of training parameters $\Gamma$, which  begs for a very large number of training samples $B$ to optimize the network.  We note that there exists a rich  literature on phase retrieval, and hence, choosing such a black-box approach for this problem would not be appropriate. Most prudent would be to develop a more mature deep learning model for this problem which incorporates of domain knowledge, while still harnessing the expressive power of deep neural network. In the following, we show how a model-aware deep architecture can be obtained for phase retrieval. 
In particular, we employ the deep unfolding approach to obtaining a class of decoder modules based on an optimization problem of the form \eqref{eq:pr2}. 

Generally speaking, finding a closed-form solution to \eqref{eq:pr2} for a given loss function is either a very difficult task or even mathematically intractable. Thus, after formalizing the objective function and the search space for a phase retrieval problem, first-order mathematical optimization techniques are utilized to tackle \eqref{eq:pr2}. Iterative optimization techniques are a popular choice for both convex and non-convex programming. Specifically, first-order methods are among the most popular and well-established iterative optimization techniques due to their low per-iteration complexity and efficiency in complex scenarios. One of the most prominent first-order optimization techniques suitable for our problem in \eqref{eq:pr2} is the projected gradient descent algorithm (PGD) \cite{sattar2020quickly,bahmani2013unifying,soltanolkotabi2017learning,lohit2019unrolled,giryes2018learned,chiang2020witchcraft}. In particular, assuming $\bz^0$ is the initial point for the algorithm, the $l$-th iteration of  PGD  to approximate the solution to $\eqref{eq:pr2}$ can be defined as follows:
\begin{align}\label{eq:PGD}
    \bz^{l+1} = \mathcal{P}_{\Omega}\left(\bz^{l} - \alpha^{l}\nabla_{\bz}\ell(\bz^l)\right),
\end{align}
where $\mathcal{P}_{\Omega}: \mathds{C}^n\mapsto \Omega\subseteq\mathds{C}^m$ denotes a mapping function of the vector argument to the feasible set $\Omega$, $\nabla_{\bz}\ell(\bz^l)$ denotes the gradient of the objective function obtained at the point $\bz^{l}$, and $\alpha^l$ represents the step-size of the PGD algorithm at the $l$-th iteration. We note that first-order methods generally suffer from a slow speed of convergence and predicting the number of iterations they require for convergence is a difficult task. As a result, they are not ordinarily recommended for real-time signal processing applications. A sensible approach to circumvent this issue is to fix the total number of iterations of such algorithms (e.g.,  $L$), and to follow up with a proper choice of the parameters for each iteration (i.e., the step-sizes) that results in the best improvement in the  objective function, while allowing only $L$ iterations. 

Let $\bg_{\bgamma_l}: \mathds{C}^n\mapsto \mathds{C}^n$, be a  parameterized mapping operator defined as
\begin{align}\label{eq:UPR0}
    \bg_{\bgamma_1}(\bz) = \mathcal{P}_{\Omega}\left(\bz - \bG^{l}\nabla_{\bz}\ell(\bz)\right),
\end{align}
where $\bgamma_{l} = \{\bG^{l}\}$ denotes the set of parameters of the mapping function $\bg_{\bgamma_l}$, and $\bG^{l}$ denotes a positive semi-definite matrix. The above mapping can then be utilized to model various first-order optimization techniques for the problem at hand, and in particular, the considered PGD algorithm in this work. Consequently, performing $L$ iterations of the form \eqref{eq:PGD} can be  modeled as follows:
\begin{align}\label{eq:decoder}
    \mathcal{G}_{\bGamma}^{L}(\bx) = \bg_{\bgamma_L} \circ \bg_{\bgamma_{L-1}} \circ \cdots \circ \bg_{\bgamma_{1}}(\mathcal{E}_{\bA}(\bx);\bx_0),
\end{align}
where the above function corresponds to the PGD in all scenarios, be it when we have a fixed step-size $\bG^{l}= \alpha\bI$,  time-varying step-sizes  $\bG^{l} = \alpha^{l}\bI$, or the general preconditioned PGD algorithm with an arbitrary choice of $\bG^{l}\succeq 0$. 
Note that the rate of convergence for the mapping function $\mathcal{G}_{\bGamma}^L$ depends heavily on the choice of the parameter set $\bGamma$, i.e., the preconditioning matrices. These parameters are usually heuristically selected in the literature and there is no straightforward methodology to obtain the best set of parameters $\bGamma$ such that $\mathcal{G}_{\bGamma}$ results in the best improvement in the objective function for applying exactly $L$ iterations. Furthermore, under some mild conditions and for a proper choice of the step-size $\alpha$, it can be shown that as $L \rightarrow \infty$, the output of the mapping function $\mathcal{G}^{L}_{\bGamma = \alpha}$ converges to a first-order optimal point of the objective function. 

In light of the above, we now formally give a high-level definition of the considered hypothesis classes of the decoder function in accordance with \eqref{eq:decoder}. Note that for a fixed realization of the set of parameters $\bGamma$ and $L$, the realized mapping function $\mathcal{G}^{L}_{\bGamma}$ can be viewed as a problem-specific decoder function.  In particular, after the formalization of the objective function $\ell(\bx)$ for a phase retrieval problem, and with an initial point of $\bx^0$, the output of the mapping function $\hat{\bx} = \mathcal{G}_{\bGamma}^L(\mathcal{E}_{\bA}(\bx);\bx^0)$ can be considered to be an estimate of the signal of interest. Accordingly, we define the hypothesis class of possible decoder functions as follows:
\begin{align}
    \mathcal{H}_{d,L} = \{\mathcal{D}_{\bGamma}: \mathds{C}^n\mapsto\mathds{C}^n | \bGamma = \{\bgamma_i = \bG^{i}\in\mathds{S}_{+}\}_{i=1}^{L}\},
\end{align}
where $\bGamma = \{\bgamma_i = \bG^{i}\in\mathds{S}_{+}\}_{i=1}^{L}$ denotes the set of parameters of the proposed class of decoder functions, $\mathds{S}_{+}$ denotes the set of all positive semi-definite matrices, and $\mathcal{D}_{\bGamma}(\mathcal{E}_{\bA}(\bx);\bx_0) = \bg_{\bgamma_L} \circ \bg_{\bgamma_{L-1}} \circ \cdots \circ \bg_{\bgamma_{1}}(\mathcal{E}_{\bA}(\bx);\bx_0)$.  
Observe that for a fixed realization $\mathcal{G}_{\bGamma^{\star}} \in \mathcal{H}_{d,L}$, one can interpret $\mathcal{G}_{\bGamma^{\star}}$ as an $L$-layer deep neural network, whose computation dynamics at each layer mimics the behaviour of one iteration of a first-order optimization algorithm as given in \eqref{eq:PGD}. Specifically, due to the incorporation of domain knowledge in the design of the deep unfolded network, the resulting deep neural network $\mathcal{G}_{\bGamma}$ is interpretable and has far fewer training parameters as compared to its black-box counterparts. Moreover, the hypothesis class $\mathcal{H}_{d,L}$ encompasses the class of gradient method realizations with a varying preconditioning matrix. We seek to find the best set of parameters of a first-order optimization algorithm with only $L$-iterations, 
or equivalently, to find $\mathcal{G}_{\Gamma}\in\mathcal{H}_{d,L}$ such that $\|\bx - \mathcal{G}_{\Gamma}\circ\mathcal{E}_{\bA}(\bx)\|$ is minimized. In the case of the phase retrieval problem and an Euclidean metric $\mathrm{D(\cdot,\cdot)}$, we are concerned with the following optimization problem:
\begin{align}\label{eq:decodertraining}
    \underset{\mathcal{G}_{\Gamma}\in\mathcal{H}_{d,L}}{\mathrm{min}}\;\;\mathds{E}_{\bx\sim P(\bx)}\{\mathrm{D}(\bx, \mathcal{G}_{\bGamma}\circ \mathcal{E}_{\bA}(\bx))\}.
\end{align}
The above program can be re-written in the more familiar form of optimizing the set of parameters $\Gamma$ containing the weights of each layer of the considered deep neural network defined in \eqref{eq:decoder}. Note that our goal is to perform a joint design over the set of parameters of both the decoder and encoder module which we address in the next part. However, for a fixed $\bA$, the learning of the deep decoder module with respect to a realization of the encoder function can be expressed as the following reformulation of \eqref{eq:decodertraining}:
\begin{align}
    \underset{\bGamma=\{\bG^i\in\mathds{S}_+\}_i}{\mathrm{min}}\;\;\mathds{E}_{\bx\sim P(\bx)}\{\mathrm{D}(\bx, \mathcal{G}_{\bGamma}\circ \mathcal{E}_{\bA}(\bx))\},
\end{align}
where the above optimization problem formulation corresponds to training an $L$-layer deep neural network $\mathcal{G}_{\bGamma}$ with a task-specific architecture as defined in \eqref{eq:decoder}, and the input to such a model-based deep architecture is given by a realization of the encoder module. In the following, we build upon the aforementioned ideas and propose two model-based deep architectures as decoder modules for both the conventional phase retrieval problem and a sparse phase retrieval settings. Specifically, we consider the unfolding of iteration of two well-established signal processing techniques for the phase retrieval problem to derive the considered decoder modules.

\vspace{0.2cm}
$\bullet$ \textbf{\emph{UPR-SPARTA: Sparse Phase Retrieval}}.
We now present the structure of the proposed UPR-SPARTA deep decoder module for the problem of sparse phase retrieval. As mentioned earlier, one particular area of interest in phase retrieval is when the signal of interest is  sparse in nature.  The applications of  sparse phase retrieval  emerge across different fields of engineering, and are seen especially in domains such as imaging \cite{kim}. Several algorithms and frameworks readily exist for sparse phase retrieval. The most notable one is the sparse truncated amplitude flow (SPARTA) methodology which seeks to recover a $s$-sparse signal from its phase-less measurements \cite{SPARTA}. Inspired by SPARTA, the following discussion is aimed at providing the iterations of a first-order gradient-based optimizer that lays the groundwork for the proposed UPR-SPARTA model-based decoder module.



 We begin our work by formalizing the problem in a non-convex form similar to \eqref{eq:pr2}. Let $\mathcal{E}_{\bA}(\bx)$ represent the encoder module providing the phase-less measurements of an underlying $s$-sparse signal $\bx$, i.e., $\bx \in \Omega:=\{\bx\in\mathds{C}^n|\|\bx\|_0 = s \}$. Then, decoding the signal $\bx$ from the measurement vector $\mathcal{E}_{\bA}(\bx)$ can be formulated as the following non-convex optimization problem: 
\begin{align}\label{eq:spartaopt}
    \underset{\bz}{\mathrm{min}}\;\;\ell(\bz;\mathcal{E}_{\bA}(\bx))\equiv\frac{1}{m}\|\mathcal{E}_{\bA}(\bx) - \mathcal{E}_{\bA}(\bz)\|_2^2,\;\;\mathrm{s.t.}\;\;\|\bz\|_0 = s.
\end{align}
Note that not only both the objective function and the constraint in the above  are non-convex, the problem is deemed to be NP-hard in its general form. The SPARTA algorithm \cite{SPARTA} was proposed  to efficiently approximate the solution to \eqref{eq:spartaopt}. In particular, the SPARTA solver works through a two-stage process. The first stage is concerned with a sparse orthogonality-promoting initialization and employs iterations based on truncated gradients. Specifically, the initialization utilizes power iterations on the estimated support of the sparse signal to solve a PCA-type problem. The second stage then makes use of truncated gradients for thresholding iterations, except for $s$ signal elements with the largest magnitudes. 
Starting from an initial guess $\bz_0$, the SPARTA algorithm is tasked with approximating the solution to \eqref{eq:spartaopt} in an iterative manner, via update equations as follows: 
\begin{align}
    \label{eq:refinedgrad}
    \bz^{l+1} = \calH_s\left(\bz^l - \alpha\nabla_{\bz}\ell_{\mathrm{tr}}(\mathcal{E}_{\bA}(\bx),\bz^l)\right),
\end{align}
where the above iterations can be viewed as performing the gradient descent algorithm with a per-iteration step-size $\alpha$, on the truncated loss function objective $\ell_{\mathrm{tr}}$ whose gradient is defined as:
\begin{align}
    \nabla_{\bz}\ell_{\mathrm{tr}}(\mathcal{E}_{\bA}(\bx), \bz)\triangleq \sum_{i \in \calI_{l+1}} \left( \ba_i^T \bz - [\mathcal{E}_{\bA}(\bx)]_i \frac{\ba_i^T \bz}{|\ba_i^T \bz|} \right) \ba_i,
\end{align}
and where $\calI_{l+1} = \{1 \leq i \leq m | | \ba_i^T \bz^l | \geq [\mathcal{E}_{\bA}(\bx)]_i/ (1+\tau) \}$, $\tau$ represents the truncation threshold, $\calH_s(\bz): \mathds{R}^{m} \mapsto \mathds{R}^{m}$ sets all entries of $\bu$ to zero except for the $s$ entries with the largest magnitudes. In particular, $\bz^0$ will be initialized as $\sqrt{\sum_{i=1}^m [\mathcal{E}_{\bA}(\bx)]_i^2/m} \hat{\bz}^0$ where $\hat{\bz}^0 \in \mathds{R}^{n}$ is created by placing zeros in $\hat{\bz}_{\hat{\calS}}^0$ where the indices are not in $\hat{\calS}$. The principal eigenvector $\hat{\bz}_{\hat{\calS}}^0 \in \mathds{R}^{s}$ is determined by performing power method iterations on the matrix:
\begin{eqnarray}
    \label{eq:eigen}
    \mathbf{\Lambda} \triangleq \frac{1}{|\calI^0|} \sum_{j \in \calI^0} \left( \frac{\ba_{i,\hat{\calS}} \ba_{i,\hat{\calS}}^T}{ \| \ba_{i,\hat{\calS}} \|_2^2} \right).
\end{eqnarray}


With the structure of the iterations of SPARTA provided in \eqref{eq:refinedgrad} and the structure provided in \eqref{eq:PGD} in mind, we can now easily derive the corresponding deep architecture by introducing a preconditioning matrix $\bG^i$ at each iteration in lieu of $\alpha$, and cast it as a trainable parameter. Ergo, UPR-SPARTA can be presented via the mathematical structure of the proposed UPR architecture in the previous part, as we define the computational dynamics of the $l$-th layer of the UPR-SPARTA architecture according to \eqref{eq:UPR0}. Furthermore, the dynamics of the overall network with $L$ layers will be the same as \eqref{eq:decoder} and the primary trainable parameters for the proposed decoder module become $\bGamma = \{\bG_0,\bG_1,\cdots,\bG_{L-1}\}$, as described in the previous section. 

For the sake of completeness, the inner computation dynamics of UPR-SPARTA is summarized below: 

\vspace{.09cm}

\begin{oframed}
\vspace{-0.3cm}
\label{UPR-SPARTA}
\noindent \\\textbf{UPR-SPARTA Computation Dynamics:} \\
Initialize \mbox{$\bz_0$} according to the discussion above \eqref{eq:eigen}. \\ Every layer $i\in\{0,1,\ldots,L-1\}$ is tuned to compute: 
\begin{equation} 
\bg_{\bgamma_i}(\bz) = \calH_s \left( \bz - \bG^i \sum_{j \in \calI_{k+1}} \bu_j \ba_j \right) \label{eq:proposedSPARTA},
\end{equation}
and
\begin{equation}\label{eq:uprSPARTA}
    \bu_j = \left( \ba_j^T \bz - [\mathcal{E}_{\bA}(\bx)]_j \frac{\ba_j^T \bz}{|\ba_j^T \bz|} \right),
\end{equation}
where $\bz$ is the input to the $l$-th layer, and $\bgamma_i = \{\bG_i \in \mathds{S}_+\}$.
The overall dynamics of UPR-SPARTA is given by:
\begin{align}
 \mathcal{G}_{\bGamma}^{L}(\mathcal{E}_{\bA}(\bx);\bz_0) \!=\! \bg_{\gamma_{L\!-\!1}} \circ \bg_{\bgamma_{L\!-\!2}} \circ \cdots \circ \bg_{\bgamma_{0}}(\mathcal{E}_{\bA}(\bx);\bz_0).\label{eq:modSPARTA}
\end{align}
\vspace{-0.2cm}
\end{oframed}


\vspace{0.2cm}
$\bullet$ \textbf{\emph{UPR-IRWF: Conventional Phase Retrieval.}} We now consider the conventional phase retrieval problem, and similar to our previous approach, consider the unfolding of the IRWF algorithm for such problems. In addition, we propose the UPR-IRWF deep architecture as a decoder module to perform the signal recovery task for a conventional phase retrieval problem. 
WF based algorithms have had an unparalleled influence on the world of phase retrieval and IRWF is no exception. In fact, IRWF's performance made it of particular interest to further explore. Specifically, as an immediate extension from RWF, IRWF was developed to tackle the following optimization problem for the recovery purposes:
\begin{align}
    \label{eq:opt}
    \underset{\bz\in\mathds{C}^m}{\min}\ell(\mathcal{E}_{\bA}(\bx); \bz)\equiv\frac{1}{2m}\Big\|\mathcal{E}_{\bA}(\bx) - \mathcal{E}_{\bA}(\bz)\Big\|_2^2.
\end{align}
The iterations of the IRWF algorithm for finding the critical points of the non-convex problem in \eqref{eq:opt} can be simply explained as follows. Starting from a proper initial point $\bz^0$ (more on this below), the IRWF algorithm generates a sequence of points $\{\bz^0, \bz^1, \bz^2, \cdots\}$ according to the following update rule:
\begin{eqnarray}
    \bz^{k+1} = \bz^{k} - \alpha\nabla_{\bz} \ell(\mathcal{E}_{\bA}(\bx); \bz),
\end{eqnarray}
where $\alpha$ is some positive step-size and the gradient of the objective function is given by:
\begin{eqnarray}
    \label{eq:iteartions}
    \nabla_{\bz}\ell(\mathcal{E}_{\bA}(\bx); \bz) = \bA^{H}\left( \bA\bz - \mathcal{E}_{\bA}(\bx)\odot\mathrm{Ph}(\bA\bz)\right),
\end{eqnarray}
where the function $\mathrm{Ph}(\bz)$ is applied element-wise and captures the phase of the vector argument; e.g., for real valued signals $\mathrm{Ph}(\bz) = \mathrm{sign}(\bz)$. 

For initialization, \cite{RWF} implements a method different than the commonly used spectral initialization, which benefits from a lower-complexity than that of the spectral method. In particular, the starting point is initialized as $\bx_0 = \lambda_0\bz$, where $\lambda_0\approx\sqrt{\pi/2}$, and $\bz$ is the leading eigenvector of the matrix $(1/m)\sum_{i=1}^{m} [\mathcal{E}_{\bA}(\bx)]_i\ba_i\ba_i^H$.

In light of the above description, we can view the mathematical structure of the proposed UPR architecture in a similar fashion as we did for UPR-SPARTA. We define the computational dynamics of the UPR-IRWF architecture as follows:

\begin{oframed}
\vspace{-0.3cm}
\label{UPR-IRWF}
\noindent \\\textbf{UPR-IRWF Computation Dynamics:} \\
Initialize \mbox{$\bz_0$} according to the discussion below \eqref{eq:iteartions}. \\ Every layer $i\in\{0,1,\ldots,L-1\}$ is tuned to compute: 
\begin{equation} 
\bg_{\bgamma_i}(\bz) = \bz - \bG^i\bu \label{eq:proposedIRWF},
\end{equation}
and
\begin{equation}\label{eq:uprIRWF}
    \bu = \bA^{H}\left(\bA\bz - \mathcal{E}_{\bA}(\bx)\odot\mathrm{sign}(\bA\bz)\right),
\end{equation}
where $\bz$ is the input to the $l$-th layer, and $\bgamma_i = \{\bG^i \in \mathds{S}_+\}$.
The overall dynamics of UPR-IRWF is given by:
\begin{align}
 \mathcal{G}_{\bGamma}^{L}(\mathcal{E}_{\bA}(\bx);\bz^0) \!=\! \bg_{\bgamma_{L\!-\!1}} \circ \bg_{\bgamma_{L\!-\!2}} \circ \cdots \circ \bg_{\bgamma_{0}}(\mathcal{E}_{\bA}(\bx);\bz_0).\label{eq:modIRWF}
\end{align}
\vspace{-0.2cm}
\end{oframed}


\section{Numerical Results}
In this section, we investigate the performance of the proposed UPR-SPARTA and UPR-IRWF frameworks for the task of phase retrieval through various numerical experiments and compare their performance relative to their base-line state-of-the-art SPARTA \cite{SPARTA} and IRWF \cite{RWF} algorithms. The numerical experiments are conducted for real-valued Gaussian signals per convention, and are averaged over $100$ Monte-Carlo trials. The proposed deep architectures are implemented using the $\rm{PyTorch}$ library \cite{paszke2017automatic}, and the optimization (training) of the networks are carried out using the Adam stochastic optimizer \cite{kingma2014adam} with a learning rate of $10^{-4}$ and for $100$ epochs. For a fair comparison, the parameters of all competing algorithms are initialized to their suggested values as reported in \cite{SPARTA} and \cite{RWF}. As a figure of merit for evaluating the performance of the proposed methodologies, we make use of the empirical success rate (ESR) and the relative mean-square error metrics. If a signal $\bx$ is to be recovered, we define the relative MSE metric for the recovered signal $\by$ as:
\begin{align}
    \text{Relative MSE} \triangleq \frac{\mathrm{D}(\by,\bx)}{\|\bx\|_2},
\end{align}
where we declare success for the recovery trial if the estimated signal assumes a relative MSE less than $10^{-5}$. For both UPR frameworks, we consider a diagonal structure on the pre-conditioning matrices, i.e., we set $\bG_i=\bW_i\bW_i^T$ , where $\bW_i = \mathrm{Diag}(\alpha_0,\alpha_1, \cdots, \alpha_{n-1})$, and perform the training over $\bW_i$, to ensure the positive-definiteness of the resulting $\bG_i$.
\subsection{Performance of the Proposed UPR-SPARTA}
In this part, we evaluate the performance of the proposed UPR-SPARTA framework for the case of recovering a $k$-sparse real-valued Gaussian signal. For training purposes, we generate a training data-set of size $2048$ where each data point $\bx\in\mathds{R}^n$ in the data-set follows a standard Gaussian distribution whose $(n-k)$ entries   set to zero are chosen randomly and uniformly. Accordingly, a testing data-set of size $2048$ with the same setting is generated using which we evaluate the performance of the proposed UPR-SPARTA and the base-line SPARTA algorithm. For all experiments in this part, the UPR-SPARTA is implemented with $L=20$ layers (iterations) and we let the base-line SPARTA algorithm run for the same number of iterations.

We first provide the exact recovery performance of the UPR-SPARTA in terms of the ESR metric defined previously, based on conducting $100$ Monte-Carlo trials. For this purpose, we set the signal dimension to $n=100$ and the sparsity level to $k=5$, while the ratio $(m/n)$ increases from $0.1$ to $3.0$. In order to show the effectiveness of the learned parameters, we compare the performance of the proposed approach with the standard SPARTA algorithm in the following scenarios:\\
$\bullet$ \emph{Case 1}: The standard SPARTA algorithm with a randomly generated measurement matrix $\bA\sim\mathcal{N}(\mathbf{0}, \bI)$ and a fixed step-size for the iterations as reported in \cite{SPARTA}.\\
$\bullet$ \emph{Case 2}: The UPR-SPARTA algorithm with a learned measurement matrix $\bA$ and a fixed step-size whose value is similar to the previous case.\\
$\bullet$ \emph{Case 3}: The UPR-SPARTA algorithm with the learned pre-conditioning matrices $\{\bG_i\}_i$ and for a fixed randomly generated $\bA$, same as \emph{Case 1}.\\
$\bullet$ \emph{Case 4}: The UPR-SPARTA algorithm with the learned measurement matrix $\bA$ and pre-conditioning matrices $\{\bG_i\}_i$.

\vspace{3pt}

\begin{figure}
	\centering
	\includegraphics[width = 0.86\linewidth]{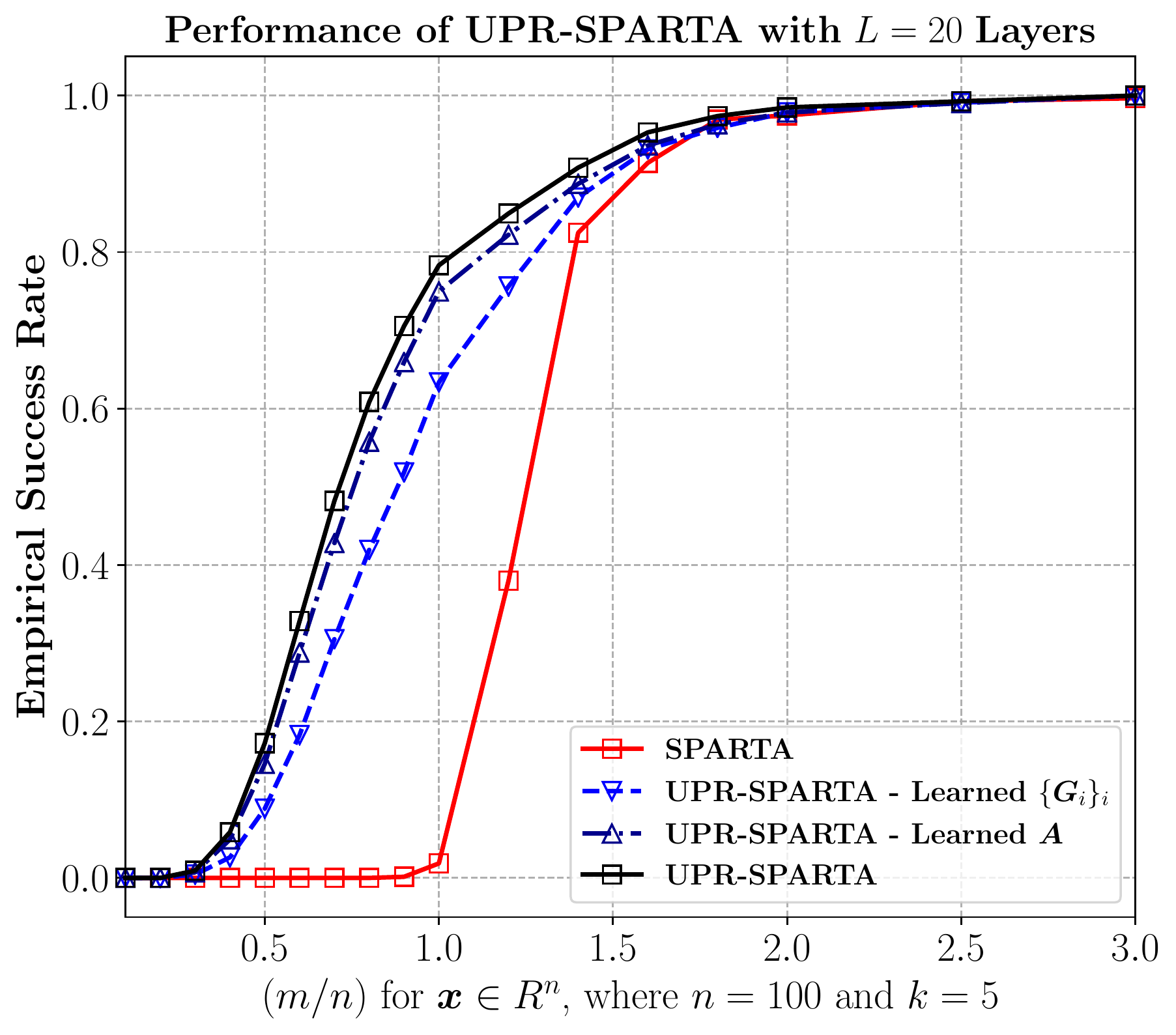}
	\caption{Empirical success rate versus the measurement to signal-length ratio ($m/n$) for $\bx\in\mathds{R}^n$ with $n=100$ and $k = 5$ nonzero entries.}
	\label{fig:ESRSparta}
\end{figure}

Fig. \ref{fig:ESRSparta} demonstrates the ESR versus the number of measurements $(m/n)$ for the above scenario. It can be observed from Fig.~\ref{fig:ESRSparta} that the proposed methodology  significantly outperforms the standard state-of-the-art SPARTA algorithm for recovering a sparse signal in all considered cases, which is further evidence for the effectiveness of the learned parameters. Moreover, a comparison between the UPR-SPARTA with learned preconditioning matrices (Case 2) and the standard SPARTA algorithm  reveals that for applications where the measurement matrix is imposed by the physics of system, the learning of the preconditioning matrices can significantly increase the performance of the recovery. On the other hand, considering a learning over the measurement matrix $\bA$ only and employing fixed step-sizes (Case 3) further indicates the effectiveness of learning task-specific measurement matrices tailored to the application at hand can significantly increase the performance of the recovery algorithm. Finally, it can be observed from Fig.~\ref{fig:ESRSparta} that the proposed UPR-SPARTA, while allowing a learning over all system parameters, outperforms all other cases and achieves a significantly higher ESR metric along all measurement ratios ($m/n$). We note that such a significant gain achieved by the proposed approach for Cases 2-4 is due to the hybrid model-based and data-driven nature of the proposed methodology.

\begin{figure}
	\centering
	\vspace{.1cm}
	\includegraphics[width = 0.86\linewidth]{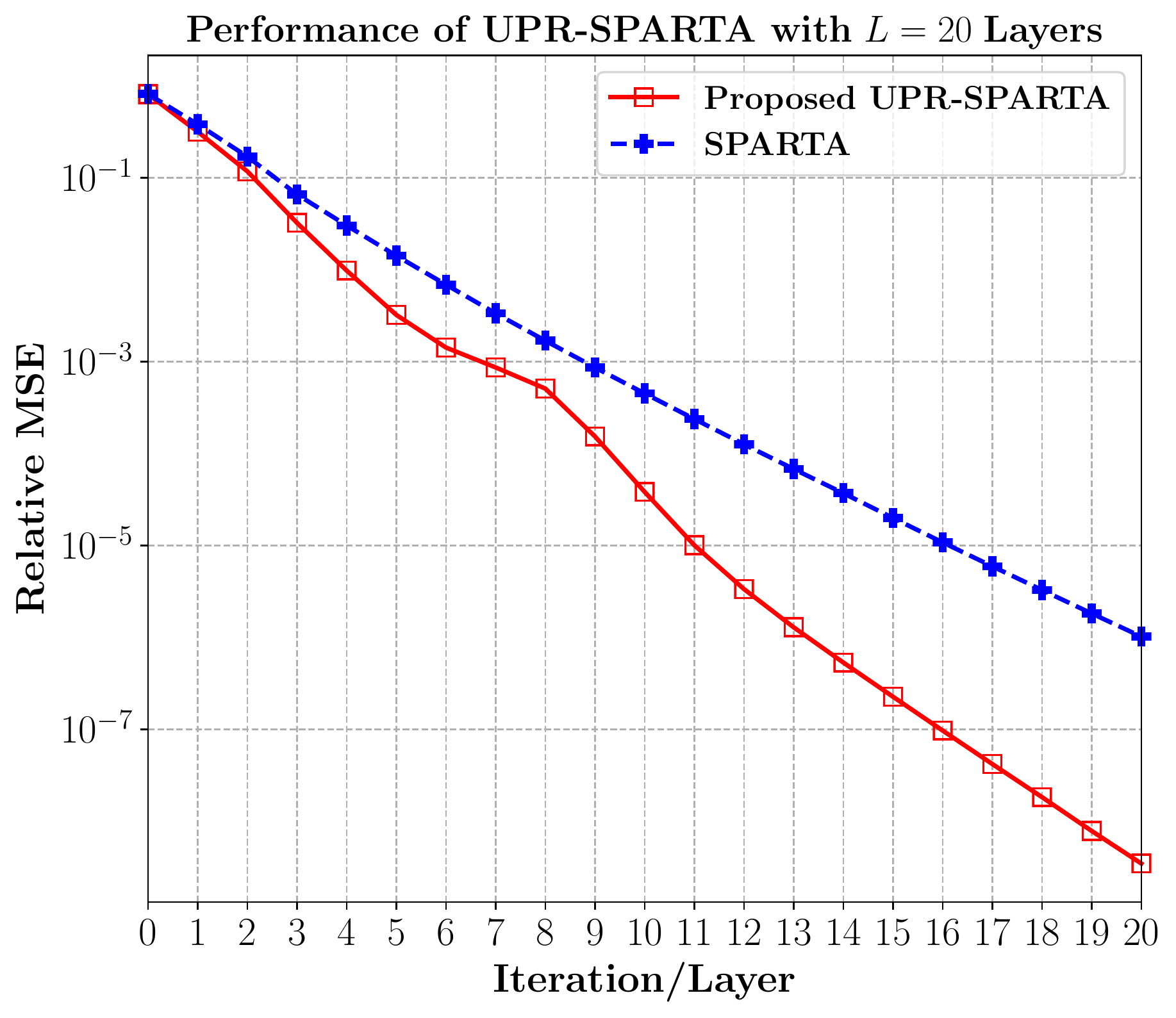}
	\caption{Convergence behavior of the proposed UPR-SPARTA as compared to the original SPARTA algorithm for the case of $\bx\in\mathds{R}^n$, where $n=m=300$.}
	\label{fig:ConvergenceSparta}
\end{figure}

We further note that learning the preconditioning matrices must lead to accelerating the convergence of the proposed methodology. In order to numerically validate this claim, we perform a per-layer analysis of the proposed UPR-SPARTA framework in terms of the achieved relative MSE. Note that such an analysis is possible due to the model-based nature of the proposed approach which results in interpretable deep architectures as opposed to conventional black-box data-driven approaches. Fig.~\ref{fig:ConvergenceSparta} demonstrates the relative MSE versus number of iterations/layers for the case of recovering a $k=5$ sparse signal $\bx\in\mathds{R}^n$, where $n=m=300$. Observing Fig. \ref{fig:ConvergenceSparta}, it can be deduced that the proposed methodology achieves a far lower relative MSE much faster than that of the standard SPARTA. 

\begin{figure}
	\centering
	\includegraphics[width = 0.86\linewidth]{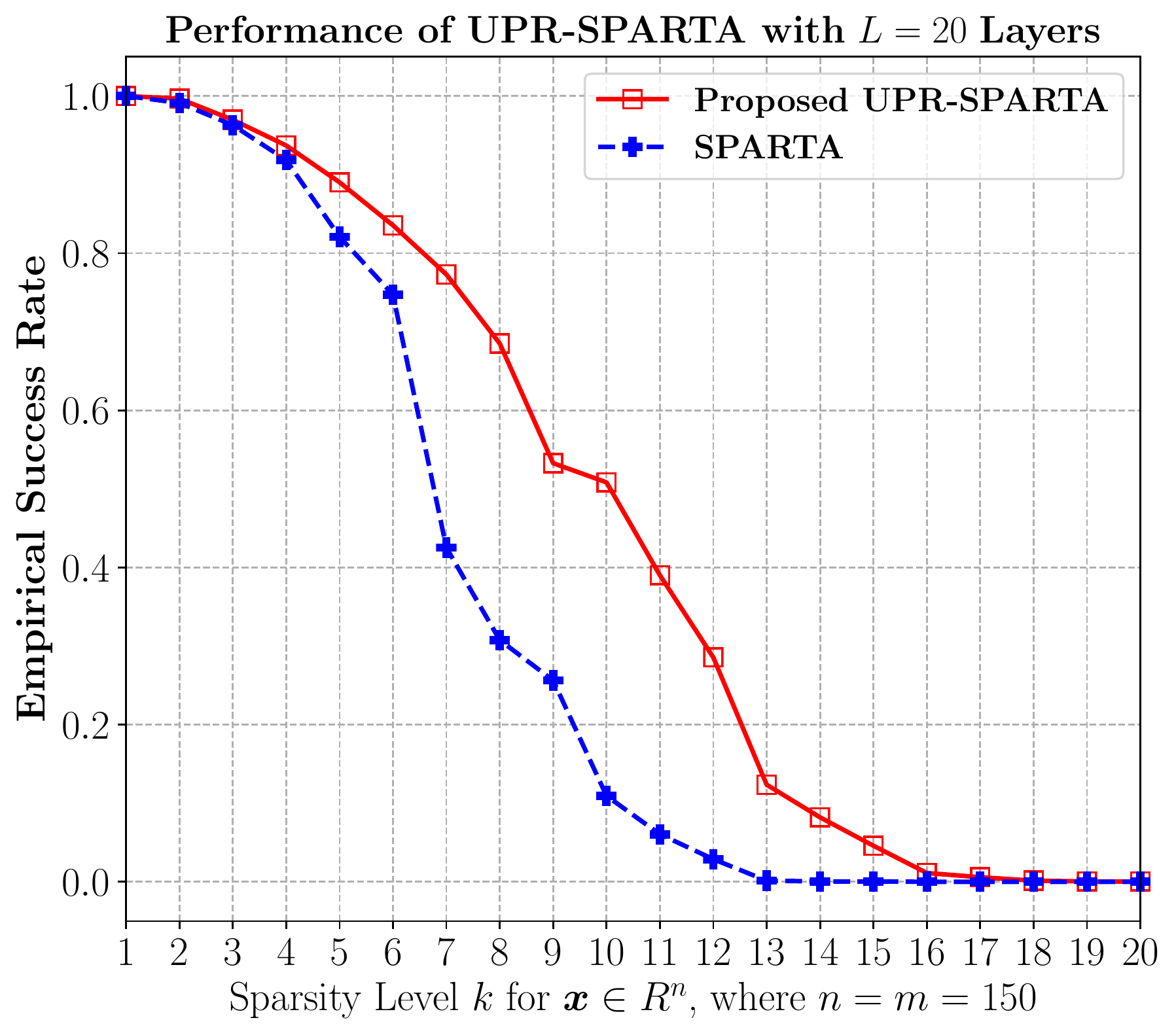}
	\caption{Empirical success rate versus the sparsity level $k$ for $\bx\in\mathds{R}^n$, where $n=m=150$.}
	\label{fig:SPARTA_Sparsity}
\end{figure}

Our final experiment in this part is an analysis of the ESR versus the sparsity level $k$, when the signal dimension and number of measurements are set to $n=m=150$. Fig.~\ref{fig:SPARTA_Sparsity} illustrates the ESR metric versus the sparsity level $k$ for the considered scenario. It can be observed from Fig. \ref{fig:SPARTA_Sparsity} that the proposed methodology has a far superior performance than that of the base-line SPARTA algorithm.

\subsection{Performance of the Proposed UPR-IRWF}
In this part, we investigate the performance of the proposed UPR-IRWF specifically tailored for conventional phase retrieval application where no prior knowledge is assumed on the underlying signal of interest (e.g., sparsity). Similar to the previous subsection, we focus on the recovering of a real-valued Gaussian signal $\bx\in\mathds{R}^n$ from the phase-less measurements. In particular, we employ the UPR-IRWF framework with $L=50$ layers (iterations), and we let the base-line IRWF algorithm run for the same number of iterations for a fair comparison.

For the first experiment, we evaluate the performance of the proposed UPR-IRWF in terms of its ESR versus number of measurements to signal-length ratio $(m/n)$ and compare its performance with the standard state-of-the-art IRWF algorithm in the following scenarios:\\
$\bullet$ \emph{Case 1}: The standard IRWF algorithm with a randomly generated measurement matrix $\bA\sim\mathcal{N}(\mathbf{0}, \bI)$ and a fixed step-size for the iterations as reported in \cite{RWF}.\\
$\bullet$ \emph{Case 2}: The UPR-IRWF algorithm with a learned measurement matrix $\bA$ and a fixed step-size whose value is similar to the previous case.\\
$\bullet$ \emph{Case 3}: The UPR-IRWF algorithm with the learned preconditioning matrices $\{\bG_i\}_i$ and for a fixed randomly generated $\bA$, same as \emph{Case 1}.\\
$\bullet$ \emph{Case 4}: The UPR-IRWF algorithm with the learned measurement matrix $\bA$ and preconditioning matrices $\{\bG_i\}_i$.

\begin{figure}
	\centering
	\includegraphics[width = 0.86\linewidth]{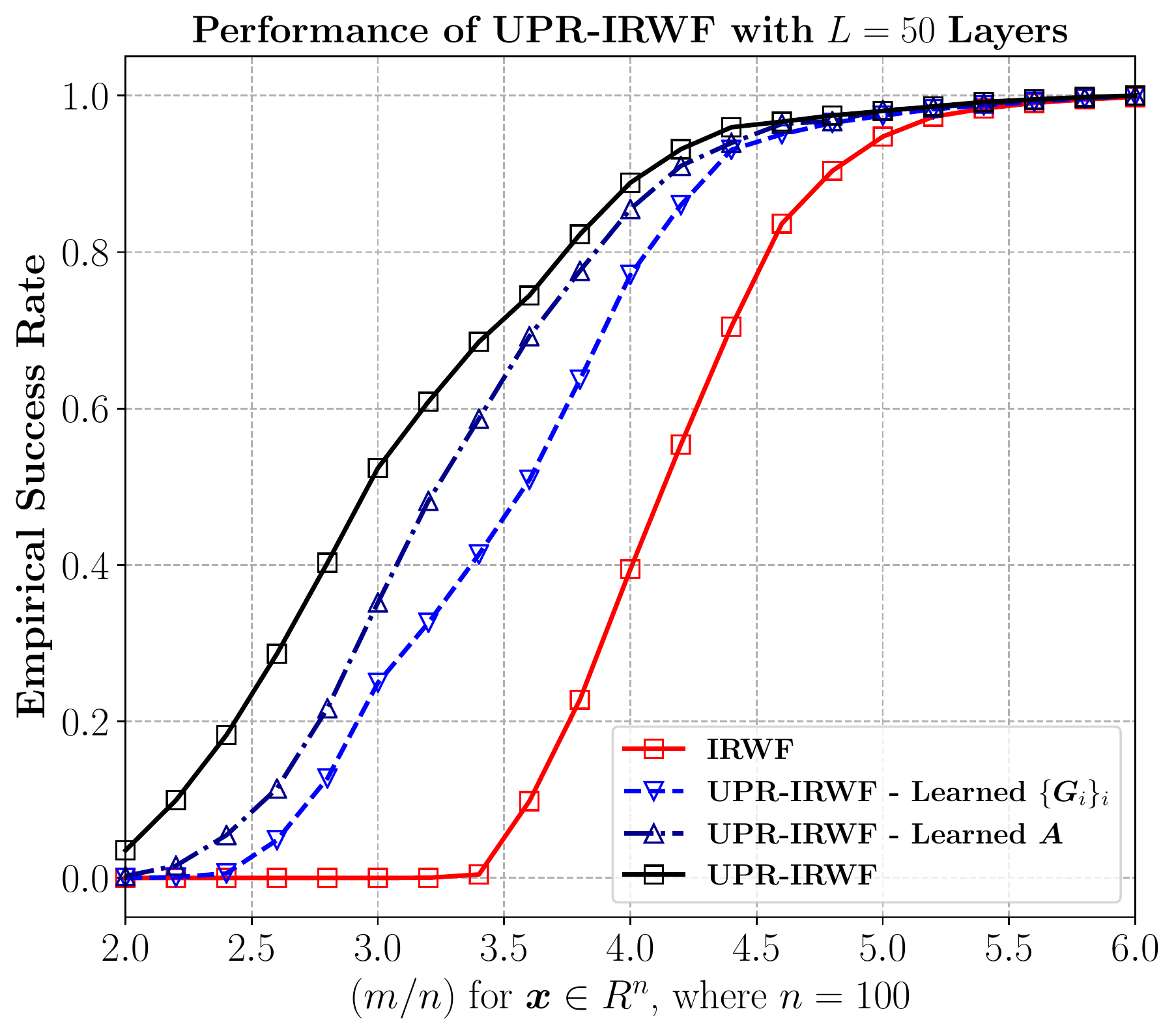}
	\caption{Empirical success rate versus the measurement to signal-length ratio ($m/n$)  for $\bx\in\mathds{R}^n$ with $n=100$.}
	\label{fig:IRWFCCDF}
\end{figure}

Fig. \ref{fig:IRWFCCDF} demonstrates the ESR versus measurement to signal-length ratio $(m/n)$ when the signal length is fixed and set to $n=100$. It can be clearly observed from Fig.~\ref{fig:IRWFCCDF} that the proposed methodology  significantly outperforms the base-line IRWF algorithm in terms of ESR in all scenarios. As for Case~3, in which we consider the scenario where the measurement matrix is fixed (i.e. imposed by the physics of the system) and is known \emph{a priori}, the learning of the preconditioning matrices significantly improves the performance of the underlying signal recovery algorithm. This supports the proposition that a judicious design of the preconditioning matrices, when the number of layers (iterations) are fixed, can indeed result in an accelerated convergence, thus immensely improving the performance. On the other hand, focusing on the Case 2 in which we only learn the measurement matrix while employing a fixed scalar step-size, further reveals the effectiveness of learning a proper task-specific data-driven measurement matrix akin to the problem  of interest. Finally, it is evident that the proposed UPR-IRWF with both learned $\bA$ and $\{\bG_i\}_i$ outperforms all other cases.

\begin{figure}[ht]
    \centering
	\hspace{-.2cm}
	\includegraphics[width =0.88\linewidth]{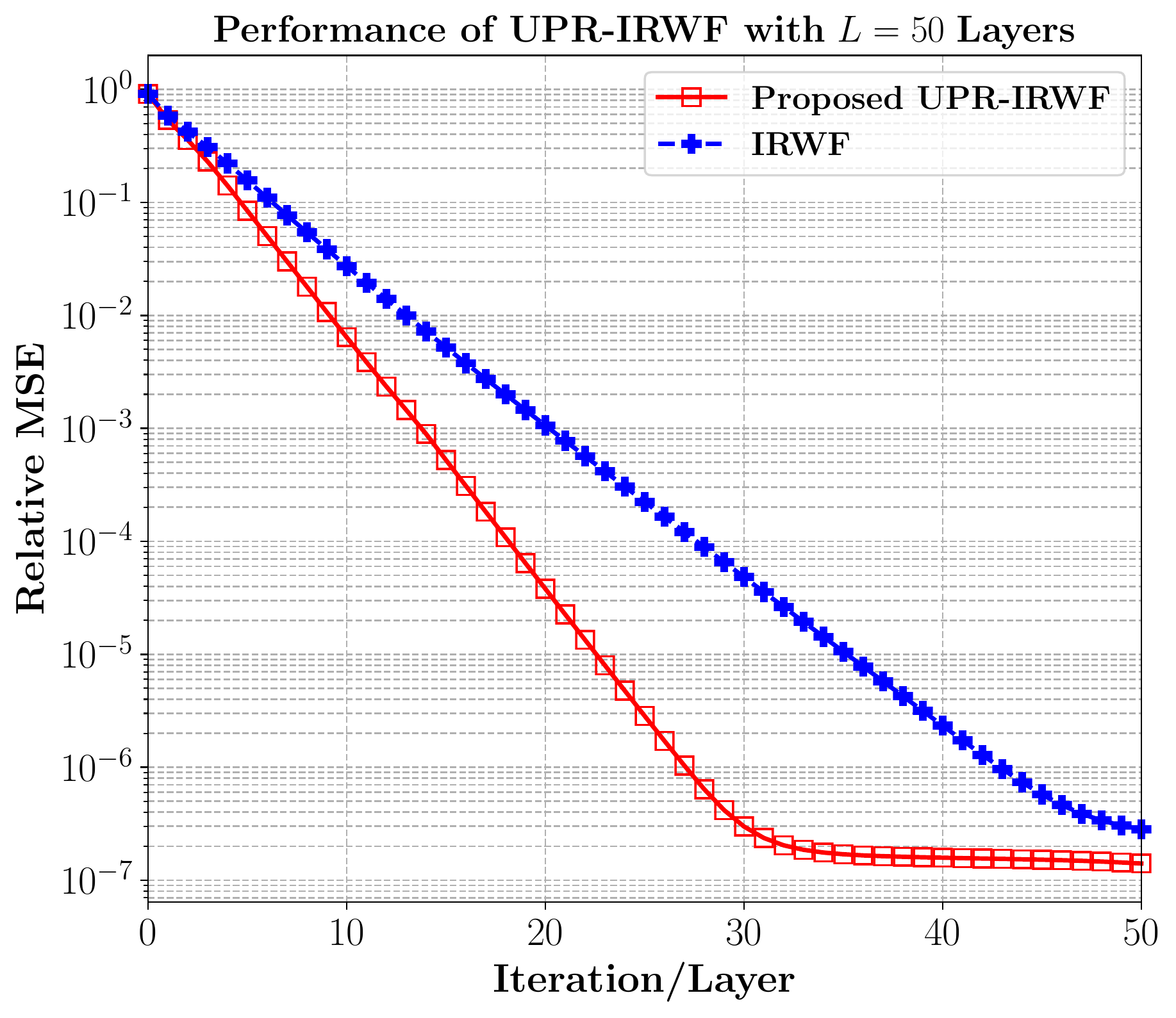}
	\caption{Convergence behavior of the proposed UPR-IRWF as compared to the original IRWF algorithm for the case of $\bx\in\mathds{R}^n$, with $(n,m) = (100, 600)$.}
	\label{fig:ConvergenceIRWF}
\end{figure}

Now, we numerically investigate the convergence properties of the proposed UPR-IRWF as compared to the standard IRWF algorithm via per-layer relative MSE analysis. We again stress that such an evaluation is possible due to the model-based nature of the proposed methodology---something that cannot be achieved when using conventional purely data-driven black-box techniques. For this experiment, we set the signal dimension and number of measurements as $(n,m) = (100,600)$. Fig.~\ref{fig:ConvergenceIRWF} demonstrates the relative MSE versus iteration (layer) number. It can be clearly observed from Fig. \ref{fig:ConvergenceIRWF} that the proposed methodology benefits from accelerated convergence properties and is able to converge to an optimal point of the objective function with as few as $L=30$ layers. This in turn reveals that the proposed methodology can be further truncated to have much fewer iterations, which will reduce the computational cost of the overall algorithm.

\nocite{weller2016robust,mukherjee2020quantization,qiao2015sparse}

\section{Conclusion}

In this paper, we proposed a new approach to the phase retrieval paper via developing hybrid model-aware and data-driven deep architectures, referred to as Unfolded Phase Retrieval (UPR). Specifically, we focused on the conventional phase retrieval problem and the sparse phase retrieval problem. 
We considered a joint design of the sensing matrix and the signal recovery algorithm, while utilizing the deep unfolding technique in the process. Such an approach allowed us to exploit the data for better accuracy and performance, and attain trusted results owing to UPR's model-based roots and the resulting interpretability. The UPR framework required less data for effective training due to having fewer parameters, and enjoyed an enhanced convergence rate. The unique capability of  UPR to allow for designing task-specific sensing matrices  further enhanced the performance of the system. 

The performance of UPR was compared with the state-of-the-art model-based phase retrieval algorithms. Our results  displayed a significant improvement in performance compared to such algorithms thanks to UPR's hybrid model-aware and data-driven nature. 

\bibliographystyle{IEEEbib}
\bibliography{refs.bib}


%








\newpage
\end{document}

%% file: symbols.tex
\usepackage{amsmath}
\usepackage{algpseudocode}
\usepackage{amsthm}
\usepackage{dsfont}
\usepackage{algorithm}
\usepackage{thmtools}
\usepackage{amssymb}

\def\ba{\boldsymbol{a}}

\def\bg{\boldsymbol{g}}

\def\bu{\boldsymbol{u}}

\def\bx{\boldsymbol{x}}
\def\by{\boldsymbol{y}}
\def\bz{\boldsymbol{z}}

\def\bA{\boldsymbol{A}}

\def\bG{\boldsymbol{G}}

\def\bI{\boldsymbol{I}}

\def\bW{\boldsymbol{W}}

\def\bgamma{\boldsymbol{\gamma}}
\def\bGamma{\boldsymbol{\Gamma}}

\def\bPhi{\boldsymbol{\Phi}}

\def\calH{\mathcal{H}}
\def\calI{\mathcal{I}}

\def\calS{\mathcal{S}}

\theoremstyle{definition}

\algnewcommand\algorithmicinput{\textbf{Input:}}
\algnewcommand\Input{\item[\algorithmicinput]}
\algnewcommand\algorithmicoutput{\textbf{Output:}}
\algnewcommand\Output{\item[\algorithmicoutput]}
\algnewcommand\algorithmicinit{\textbf{Initialize:}}
\algnewcommand\Init{\item[\algorithmicinit]}
